\pdfoutput=1

\documentclass[11pt]{article}

\usepackage{emnlp2021}

\usepackage{times}
\usepackage{latexsym}
\usepackage{graphicx}
\usepackage{xcolor}
\usepackage{listings}
\usepackage{xparse}
\NewDocumentCommand{\codeword}{v}{%
\begin{footnotesize}
\texttt{\textcolor{blue}{#1}}%
\end{footnotesize}
}

\usepackage[T1]{fontenc}
\usepackage[utf8]{inputenc}
\usepackage{microtype}

\definecolor{lmdiffRed}{RGB}{214, 96, 77}
\definecolor{lmdiffBlue}{RGB}{67, 147, 195}
\newcommand{\mone}[1]{\textcolor{lmdiffRed}{#1}}
\newcommand{\mtwo}[1]{\textcolor{lmdiffBlue}{#1}}

\newcommand{\lmdiff}[0]{\textsc{LMdiff}}

\usepackage{tabularx}
\usepackage{booktabs}
\usepackage{makecell}
\usepackage{hyperref}

\title{\lmdiff: A Visual Diff Tool to Compare Language Models}

\author{%
  Hendrik Strobelt\\
   IBM Research\\
  MIT-IBM Watson AI Lab\\
  \texttt{hendrik@strobelt.com} \\
   \And
   Benjamin Hoover \\
   IBM Research\\
   College of Computing, Georgia Tech\\
   \texttt{benjamin.hoover@ibm.com} \\
   \AND
   Arvind Satyanarayan \\
   MIT CSAIL \\
   Massachusetts Institute of Technology \\
   \texttt{arvindsatya@mit.edu} \\
   \And
   Sebastian Gehrmann \\
   Google Research \\
   Harvard University\\
   \texttt{gehrmann@google.com} \\
}

\begin{document}
\maketitle
\begin{abstract}
 While different language models are ubiquitous in NLP, it is hard to contrast their outputs and identify which contexts one can handle better than the other. To address this question, we introduce \lmdiff, a tool that visually compares probability distributions of two models that differ, e.g., through finetuning, distillation, or simply training with different parameter sizes. \lmdiff~allows the generation of hypotheses about model behavior by investigating text instances token by token and further assists in choosing these interesting text instances by identifying the most interesting phrases from large corpora. We showcase the applicability of \lmdiff~for hypothesis generation across multiple case studies. 
 A demo is available at \url{http://lmdiff.net}. 
\end{abstract}

\section{Introduction}

Interactive tools play an important role when analyzing language models and other machine learning models in natural language processing (NLP) as they enable the qualitative examination of examples and help assemble anecdotal evidence that a model exhibits a particular behavior in certain contexts. This anecdotal evidence informs hypotheses that are then rigorously studied~\citep[e.g.,][]{tenney-etal-2019-bert,belinkov-glass-2019-analysis,rogers-etal-2020-primer}. 
Many such tools exist, for example to inspect attention mechanisms~\citep{hoover-etal-2020-exbert,Vig2019AMV}, explain translations through nearest neighbors~\citep{strobelt2018seq2seq}, investigate neuron values~\citep{dalvi2019neurox,strobelt2017lstmvis}, and many more that focus on the outputs of models~\citep[e.g.,][]{cabrera2019fairvis}. 
There also exist multiple frameworks that aggregate methods employed in the initial tools to enable others to extend or combine them~\citep{pruksachatkun-etal-2020-jiant,wallace-etal-2019-allennlp,tenney-etal-2020-language}.

However, notably absent from the range of available tools are those that aim to \textit{compare} distributions produced by different models. While comparisons according to performance numbers are common practice in benchmarks~\citep{wang-etal-2018-glue,hu2020xtreme,gehrmann2021gem}, there exists only rudimentary support in existing tools for inspecting how model outputs compare for specific tasks or documents. 
Yet, this problem motivates many current studies, including questions about how models handle gendered words, whether domain transfer is easy between models, what happens during finetuning, where differences lie between models of different sizes, or how multilingual and monolingual models differ. 

\begin{figure*}[t]
    \centering
    \includegraphics[width=\linewidth]{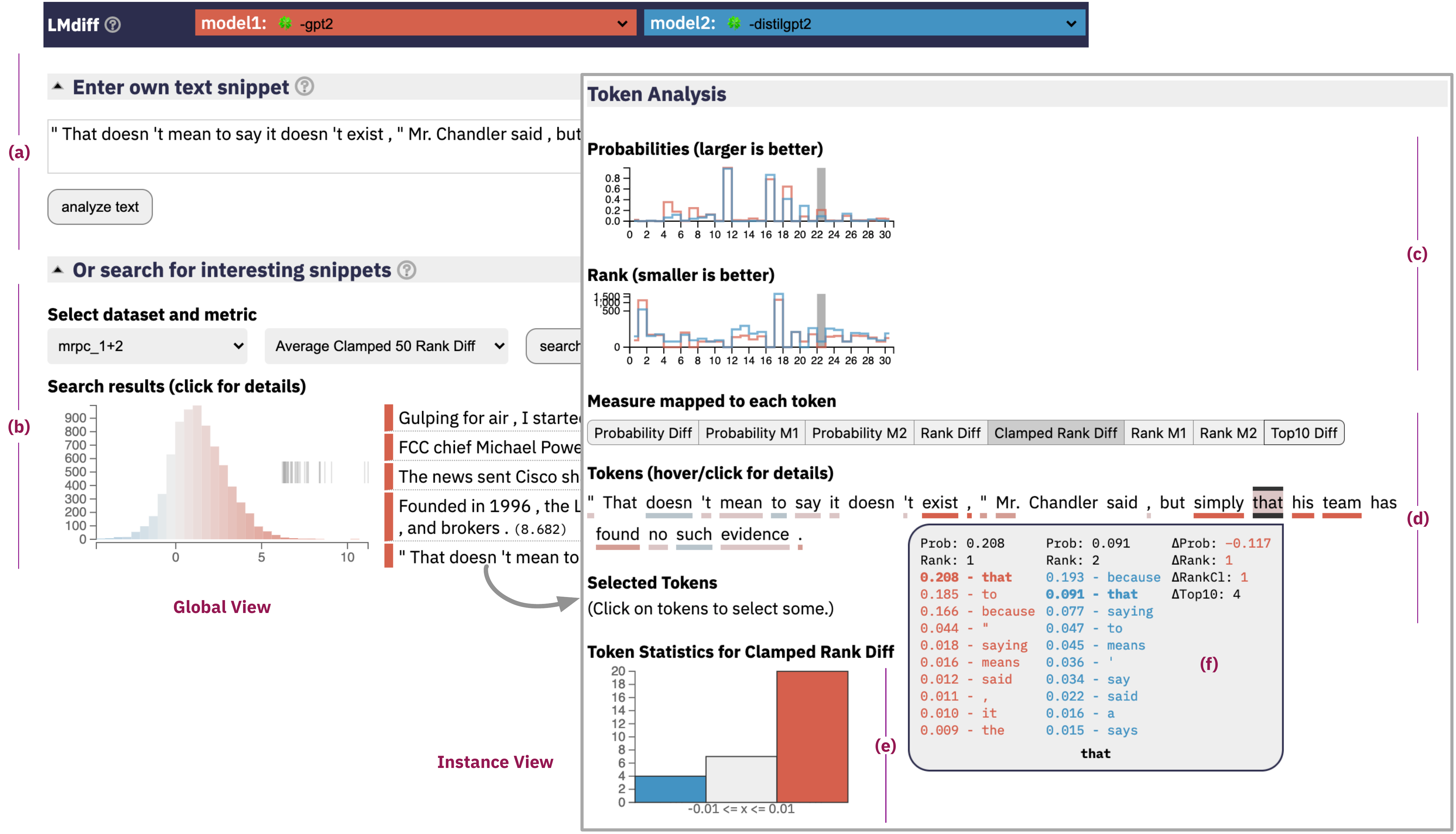}
    \caption{\lmdiff{} interface. The Global View (a,b) allows finding interesting examples which are then selected for in-depth investigation in the Instance View (c-f).}
    \label{fig:teaser}
\end{figure*}

To fill this gap, we introduce \lmdiff: an interactive tool for comparing language models by qualitatively comparing per-token likelihoods. Our design provides a \textit{global} and a \textit{local} view: In the global step, we operate on an entire corpus of texts, provide aggregate statistics across thousands of data points, and help users identify the most interesting examples. An interesting example can then be further analyzed in the local view. Fine-grained information about the model outputs for the chosen example is visualized, including the probability of each token and the difference in rank within each model's distribution. 
Similar to other visual tools, \lmdiff{} helps form hypotheses that can then be tested through rigorous statistical analyses. Across six case studies, we demonstrate how it enables an effective exploration of model differences and motivates future research. 
A deployed version of \lmdiff{} with six corpora and nine models is available at \url{http://lmdiff.net/} and its code is released at \url{https://github.com/HendrikStrobelt/LMdiff} (Apache 2.0 license) with support to easily add additional models, corpora, or evaluation metrics.

\begin{table*}[t]
\small
    \centering
    \begin{tabularx}{\textwidth}{lXl}
        \toprule
         \textbf{Dataset} &  \textbf{Description} \\ \midrule
        \makecell[tl]{WinoBias \\ \citep{zhao-etal-2018-gender} }& Collection of 3,160 sentences using different resolution systems to understand gender bias issues. We include two versions: (a) just sentence, (b) sentence with addendum (e.g., ``he refers to doctor'')\\ \midrule
        \makecell[tl]{CommonsenseQA \\ \citep{talmor-etal-2019-commonsenseqa} }& Collection of 12,102 questions with one correct answer and four distractor answers. For our use cases, we concatenate the question and the correct answer to one single string.\\ \midrule
        \makecell[tl]{MRPC \citep{dolan-etal-2004-unsupervised} } & Collection of 5,801 sentence pairs collected from newswire articles.\\ \midrule
        \makecell[tl]{GPT2-GEN \\ \citep{gpt2out} } & Collection of generated sentences from GPT-2 models. For each model the dataset contains 250K random samples (temperature 1, no truncation) and 250K samples generated with Top-K 40 truncation. We use the subset
        GPT-2-762M k40. \\ \midrule
       \makecell[tl]{Short Jokes \citep{jokes} } & Collection of 231,657 short jokes provided as Kaggle challenge for humor understanding. \\ \midrule
       \makecell[tl]{BioLang \\ \citep{liechti2017sourcedata}}& Collection of $~$12 million abstracts and captions from open access Europe PubMedCentral processed by the EMBO SourceData project\\ 

        \toprule
         \textbf{Model} & \\ \midrule
         \makecell[tl]{GPT-2 \citep{radford2019language}} & The decoder of a Transformer  trained on OpenWebText\\ 
         DistilGPT-2 & A smaller Transformer trained to replicate GPT-2 output \\ 
        GPT-2-ArXiv & GPT-2 finetuned on a large arxiv dataset \\ 
        GPT-2-ArXiv-NLP & GPT-2 finetuned only on arxiv NLP papers \\ \midrule
       \makecell[tl]{BERT-base-uncased \\ \citep{devlin2018bert}} & Masked language model with case-insensitive tokenization. \\ 
        \makecell[tl]{DistilBERT \citep{sanh2019DistilBERTAD}} & A smaller Transformer trained to replicate BERT output \\
        DistilBERT-SST-2 & distilBERT finetuned on the SST-2 \citep{socher-etal-2013-sst} dataset \\ 
        
        \midrule
        GPT-2-German & GPT-2 trained on various German texts \\
        GPT-2-German-Faust & The German GPT-2 model finetuned on Faust I \& II \\
    
        \bottomrule
    \end{tabularx}
    \caption{The default corpora and models found in the deployed version of \lmdiff. All models were taken from Huggingface's \href{https://huggingface.co/models}{model hub}. Horizontal lines group tokenization-compatible models.}
    \label{tab:data-and-models}
\end{table*}

\section{Methods}
\label{Sec:methods}
\lmdiff~compares two models $m_{\{1,2\}}$ by analyzing their probability distributions at the position of each token $\hat{X}_{1:N}$ in a specific text. 
A correct token's probability distribution $p_{\mathrm{m_j}}(X_i=\hat{X}_i | X_{1: i-1})$ is easily influenced the scaling factor $\beta$ in the function $p = \mathrm{softmax}\big(\beta x)$ used to convert logits $x$ into probabilities $p$ (though two distributions are still comparable if both use the same $\beta$). For this reason, we also include the correct token's rank in $p_{\mathrm{m_j}}(X_i |X_{1:i-1})$. 
From the probabilities and ranks, we derive eight measures of global difference (comparison over a corpus) and eight measures of local difference (comparison over an example).
The global measures are the (1) difference in rank of each token, (2) the difference in rank after clamping a rank to a maximum of 50, (3) the difference in probability of each token, and (4) the number of different tokens within the top-10 predicted tokens.\footnote{Other metrics like the KL-Divergence were omitted from the final interface since the numbers were too hard to interpret.} For each measure, we allow filtering by its average or maximum in a sequence.

To compare two models on a single example, we either directly visualize $p_{m_1}(X_i=\hat{X}_i)$, $p_{m_2}(X_i=\hat{X}_i)$, $p_{m_1}(X_i=\hat{X}_i) - p_{m_2}(X_i=\hat{X}_i)$, or the equivalent measures but focusing on the rank instead of the probability. As for the global measures, we present rank differences in both an unclamped and a clamped version. The clamped version surfaces more interesting examples; e.g., the difference between a token of rank 1 and 5 is more important than the rank difference between 44 and 60. The visual interface maps the difference to a blue-red scale (see \autoref{fig:teaser}d) and visualizations of a single model to a gray scale.

\subsection{Visual Interface}

\autoref{fig:teaser} shows the \lmdiff{} interface.
The user starts their investigation by specifying the two models $m_1$ and $m_2$ and a target text $d$. This target may either be entered into the free-text field (\ref{fig:teaser}a) or chosen from the list of suggested interesting text snippets (\ref{fig:teaser}b, see Section~\ref{Sec:candidates}). Upon selection of the text, the likelihoods, ranks, and difference metrics for $m_1$ and $m_2$ for each token of $d$ are computed.

Users can compare results using the instance view, which leverages multiple visual idioms to show aspects of the models' performance.
The step plots (\autoref{fig:teaser}c) show the absolute values for likelihoods and ranks, with color indicating the model. 
Upon selecting a distance metric, it is mapped onto the text (\ref{fig:teaser}d) using a red-white-blue diverging color scheme: white for no or minimal distance, red/blue for values in favor of a corresponding model. For instance, a token is colored blue if the rank of that token under model $m_2$ is lower than under $m_1$ or its likelihood higher. The highlighting on hover between both plots (\ref{fig:teaser}c+d) is synchronized, to help spot examples where the measures diverge. 

The histogram (\ref{fig:teaser}e) indicates the distribution of measures for the text. If the centroid of the histogram leans decidedly to one side, it indicates that one model is better at reproducing the given text (observe the shift for red in \autoref{fig:teaser}e). The token detail view (\ref{fig:teaser}f), shows all difference measures for a selected token and allows for a direct comparison of the top-k predictions for each model at the token position. E.g., in  \autoref{fig:teaser}f, the token ``that'' has rank 1 in model $m_1$ but rank 2 in $m_2$. Clicking tokens makes the detail view for those tokens stick to the bottom of the page to enable investigations of multiple tokens in the same sequence.  

\subsection{Finding Interesting Candidates }
\label{Sec:candidates}
To facilitate searching for interesting texts, we extract examples from a large corpus of texts for which the two models differ the most. The corpus is prepared via an offline preprocessing step in which the differences between the models are scored according to the methods outlined above. Each example is compared using different aggregation methods, like averaging, finding the median, the upper quartile, or the top-k of differences in likelihoods, ranks, and clamped ranks. The 50 highest-ranking text snippets for each measure are considered as interesting.
The interface (\autoref{fig:teaser}b) shows a histogram of the distribution of a measure over the entire corpus and indicates through black stripes where interesting outlier samples are located fall on the histogram. That way, users can get an overview of how the two models compare across the corpus while also being able to view the most interesting samples. 

\section{Supported Data and Models}

The deployed version of \lmdiff{} currently supports six datasets and nine models, detailed in Table~\ref{tab:data-and-models}. All pretrained models were taken from Huggingface's model hub\footnote{\url{https://huggingface.co/models}}. Section~\ref{sec:design} explains how to use \lmdiff{} with many more custom models and datasets.

\section{Case Studies}
\label{sec:case_studies}
 As discussed above, this tool aims to generate hypotheses by discovering anecdotal evidence for certain model behavior. It will not be able to give definite proofs for discovered hypotheses, which should instead be explored more in-depth in follow-up studies. As such, in this section, we provide examples of new kinds of questions that \lmdiff~helps investigate and explore further questions inspired by past findings.

\begin{figure}[t]
    \centering
    \includegraphics[width=\columnwidth]{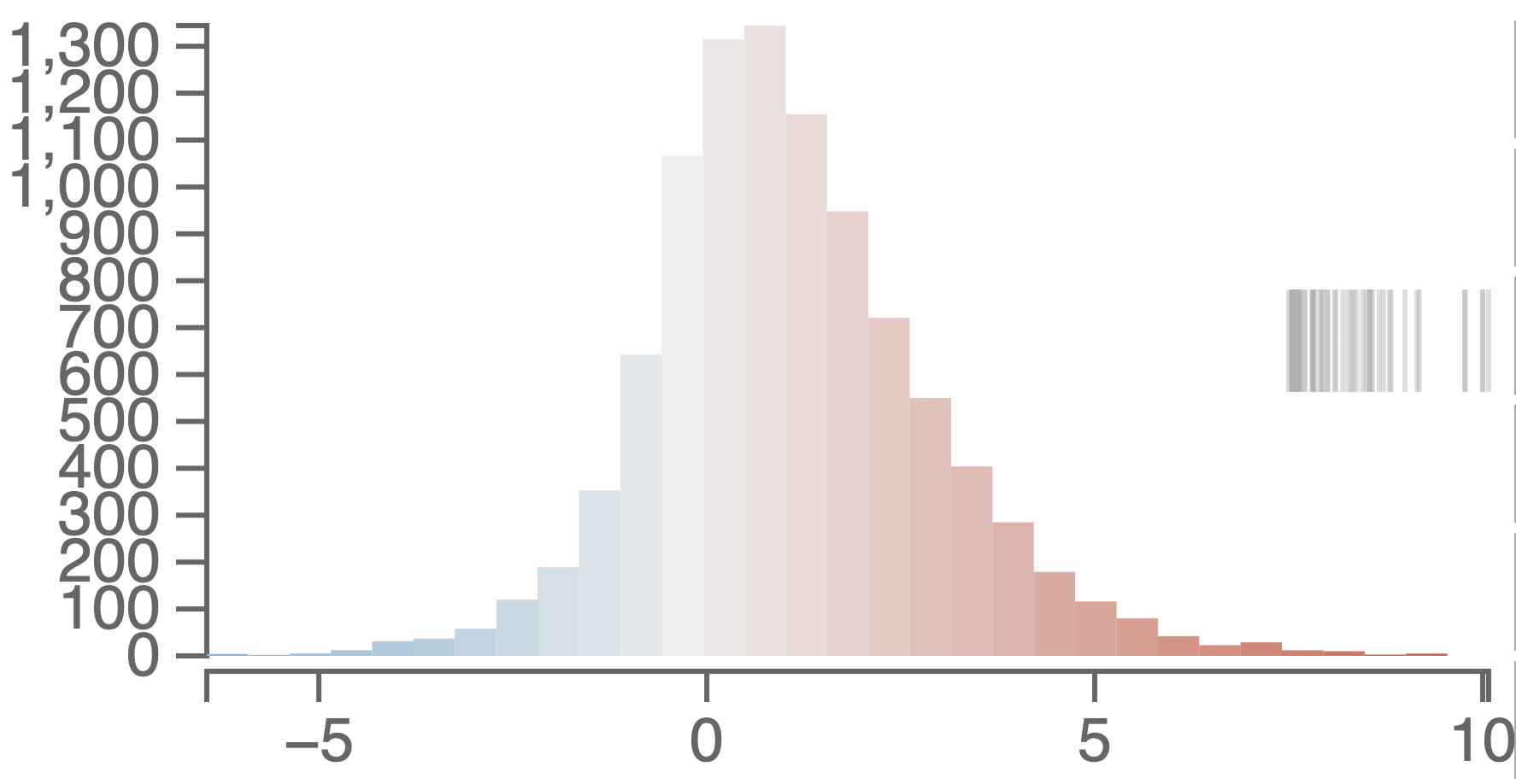}
\caption{The global view on the CommonsenseQA dataset when comparing \mone{GPT-2} and \mtwo{DistilGPT-2}. The histogram depicts the distribution of a specific measure (Average Clamped 50 Rank) over the reference corpus. The short black lines depict the values of the 20 highest values.}
    \label{fig:commonsense-overview}
\end{figure}

\subsection{Which model is better at commonsense reasoning?}

\begin{figure}[t]
    \centering
    \includegraphics[width=\columnwidth]{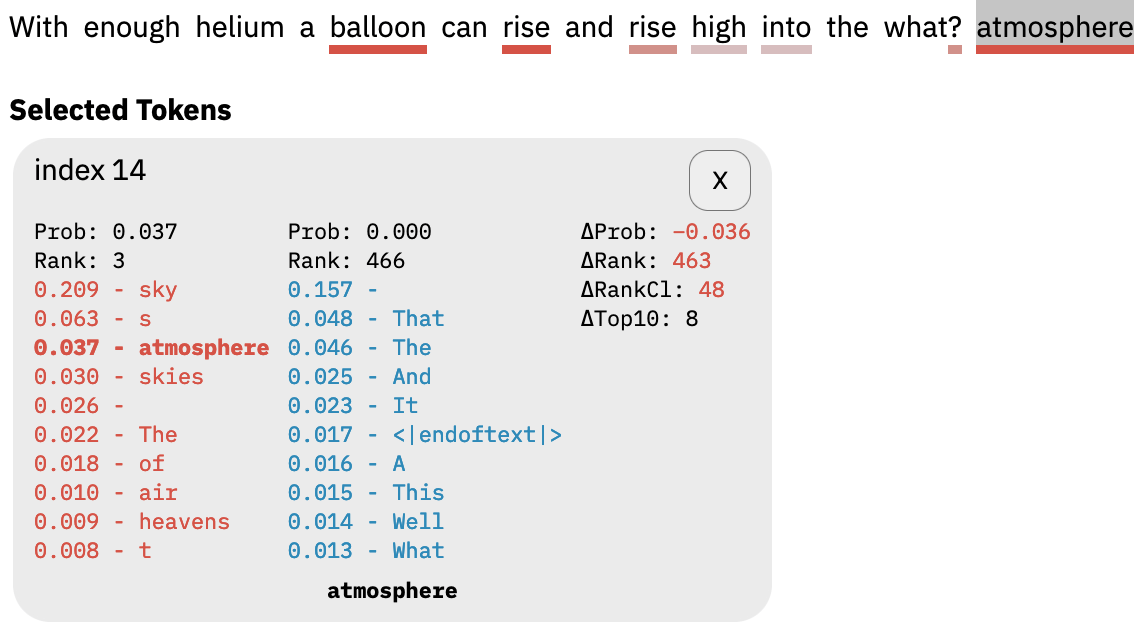}
\caption{A commonsenseQA example in which \mone{GPT-2} performs much better than \mtwo{DistilGPT-2}. Showing Clamped Rank difference. 
}
    \label{fig:commonsense-example}
\end{figure}

Prompt-based approaches have become a popular way to test whether a model can perform a task~\citep{DBLP:conf/nips/BrownMRSKDNSSAA20}. A relevant question to this is whether models can perform tasks that require memorization of commonsense knowledge (e.g., the name of the company that develops Windows, or the colors of the US flag)~\citep{DBLP:journals/tacl/JiangXAN20}. For our case study, we format the CommonsenseQA~\citep{talmor-etal-2019-commonsenseqa} dataset to follow a ``Question? Answer'' schema, such that we can compare the probability of the answer under different models. Comparing GPT-2 (red) and its distilled variant DistilGPT-2 (blue), we can observe in Figure~\ref{fig:commonsense-overview} that overall, GPT-2 performs much better on the task, commonly ranking the correct answer between 1 and 5 ranks higher in its distribution. An interesting example shown in Figure~\ref{fig:commonsense-example} paints a particularly grim picture for DistilGPT-2 --- while the standard model ranks the correct answer third, the distilled variant ranks it 466th. This leads to the questions of why this bit of knowledge (and those of other outliers) was squashed in the distillation process, whether there is commonality between the forgotten knowledge, and it motivates the development of methods that prevent this from happening. 

\begin{figure}[t]
    \centering
    \includegraphics[width=\columnwidth]{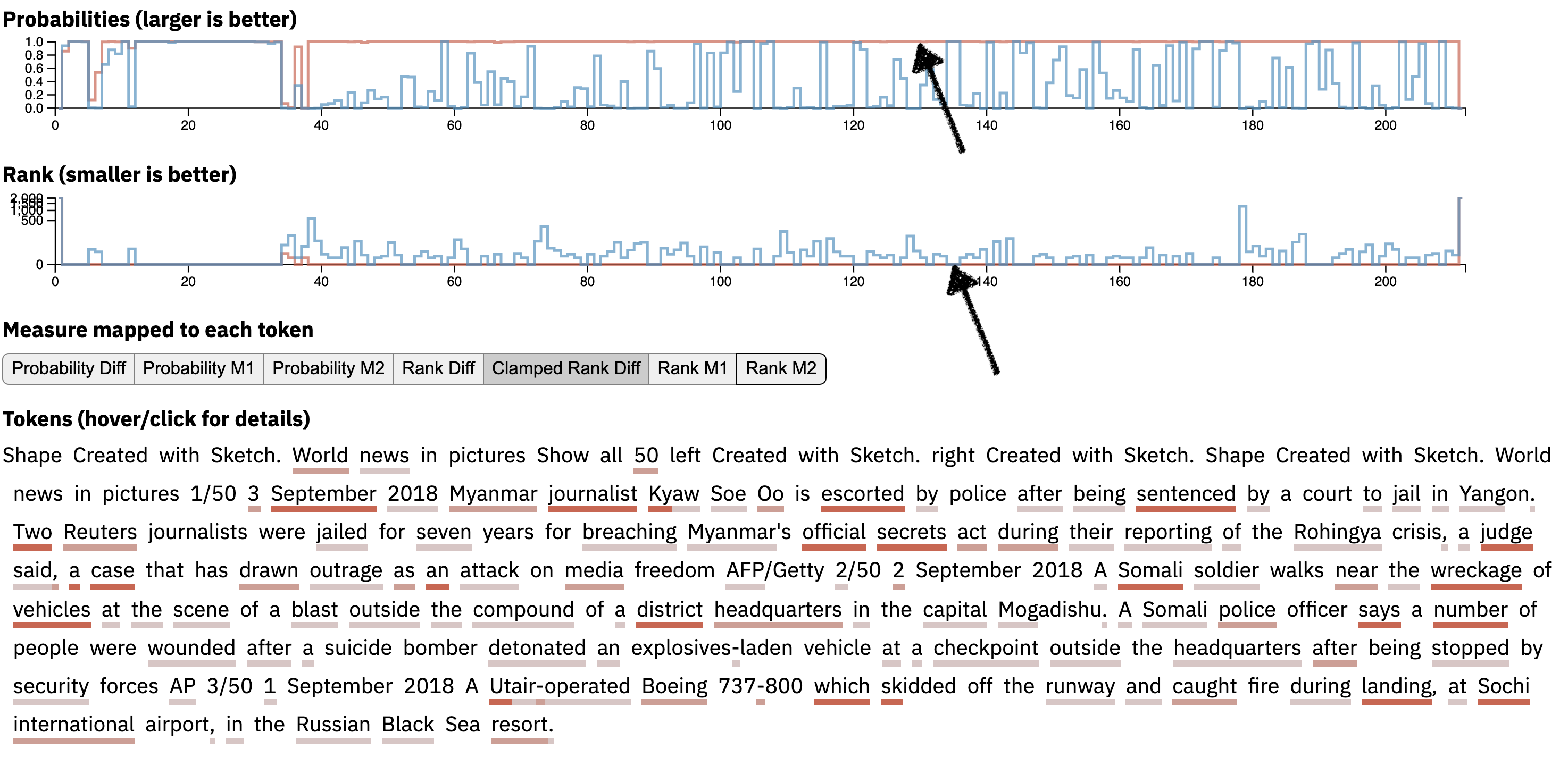}
\caption{Comparing \mone{GPT-2} vs \mtwo{DistilGPT-2} on GPT-2 generated text shows that it is easy to spot which model produced it. 
}
    \label{fig:gltr}
\end{figure}

\subsection{Which model produced a text?}
\label{sec:gltr}

Prior work has investigated different ways to detect whether a text was generated by a model or written by a human, either by training classifiers on samples from a model~\citep{DBLP:conf/nips/ZellersHRBFRC19,DBLP:conf/nips/BrownMRSKDNSSAA20} or directly using a models probability distribution~\citep{gehrmann-etal-2019-gltr}. A core insight from these works was that search algorithms (beam search, top-k sampling, etc.) tend to sample from the head of a models' distribution. That means that it is visually easy to detect if a model generated a text. With \lmdiff, we extend upon this insight to point to \textit{which} model generated a text --- if a model generated a text, the text should be consistently more likely under that model than under other similar models. While our tool does not allow us to test this hypothesis at scale, we can find clear anecdotal evidence shown in Figure~\ref{fig:gltr}. In the figure, we compare the probabilities of GPT-2 and DistilGPT-2 on a sample of GPT-2 generated text. We observe the consistent pattern that GPT-2 assigns an equal or higher likelihood to almost every token in the text.

\begin{figure}[t]
    \centering
    \includegraphics[width=\linewidth]{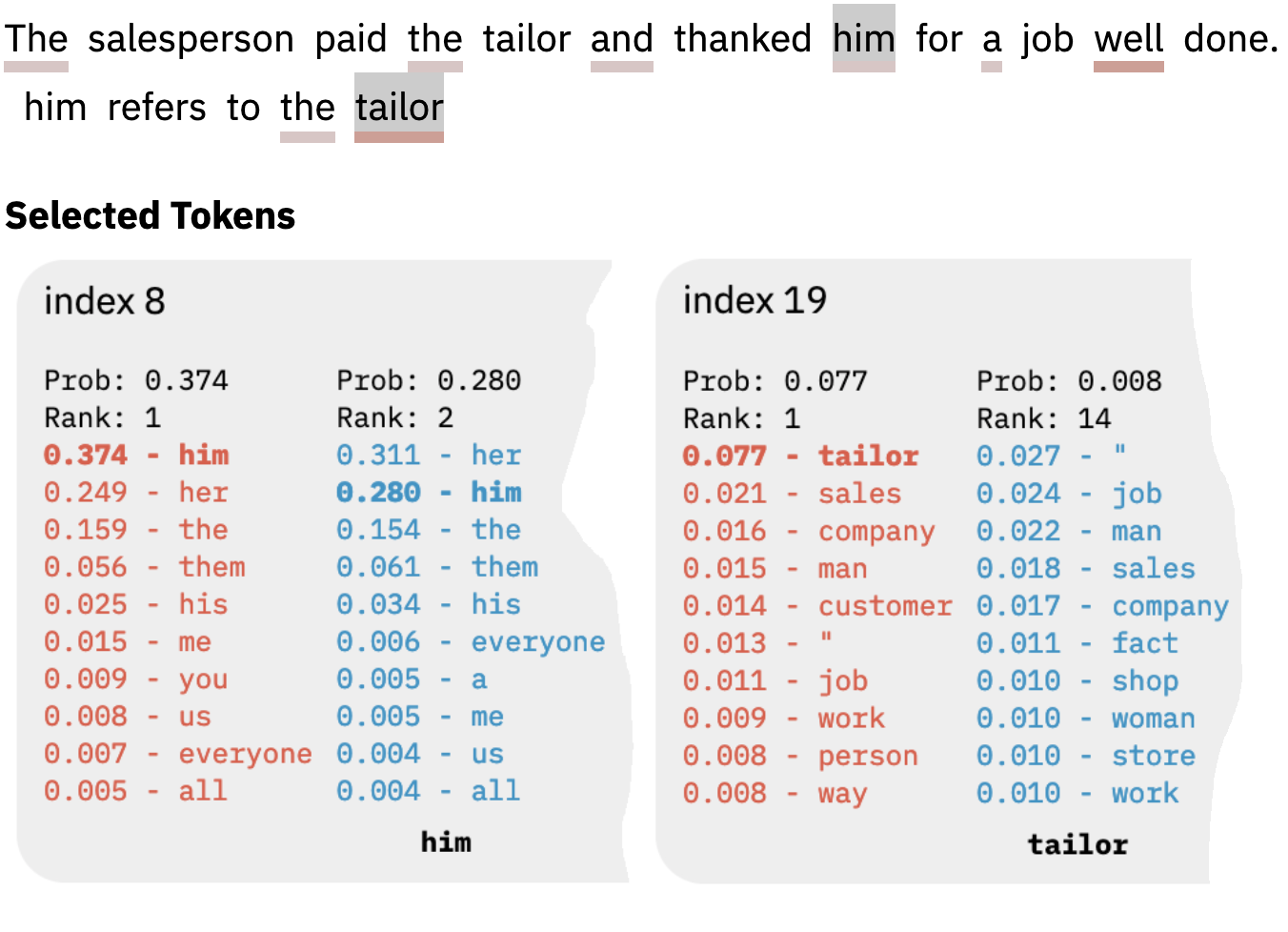}
\caption{Winobias example with addendum for \mone{GPT-2} vs \mtwo{DistilGPT-2} showing Clamped Rank difference. Interesting since him/her probability rank switches between models and only distil fails at the addendum task.}
    \label{fig:winobias}
\end{figure}

\subsection{Which model is more prone to be overconfident in coreference?}
We next investigate whether one model has learned spurious correlations in coreference tasks, using our augmented version of the WinoBias dataset~\citep{zhao-etal-2018-gender}. Since we are comparing language models, we modified the text to add the string ``\textit{[pronoun]} refers to the \textit{[profession]}''. We can then use the detail view to look at the probabilities of the pronoun in the original sentence and the probability of the disambiguating mention of the profession. In our example (Figure~\ref{fig:winobias}), we again compare GPT-2 (red) and DistilGPT-2 (blue). Curiously, the distillation process flipped the order of the predicted pronouns ``him'' and ``her''. Moreover, DistilGPT-2 fails to complete the second sentence while GPT-2 successfully predicts ``Tailor'' as the most probable continuation, indicating that DistilGPT-2 did not strongly associate the pronoun with the profession.  
This case study motivates further investigation of cases where distillation does not maintain the expected ranking of continuations. A similar effect has previously been detected in distillation processes for computer vision models~\citep{hooker2020characterising}.

\subsection{What predictions are affected the most by finetuning?}
\label{sec:domain}
Other, more open-ended, qualitative comparisons that are enabled through \lmdiff{} aim to understand how a model changes when it is finetuned on a specific task or documents from a specific domain. The finetuning process can impact prediction both in the downstream domain and in not anticipated, unrelated other domains.

\begin{figure}[ht]
    \centering
    \includegraphics[width=\columnwidth]{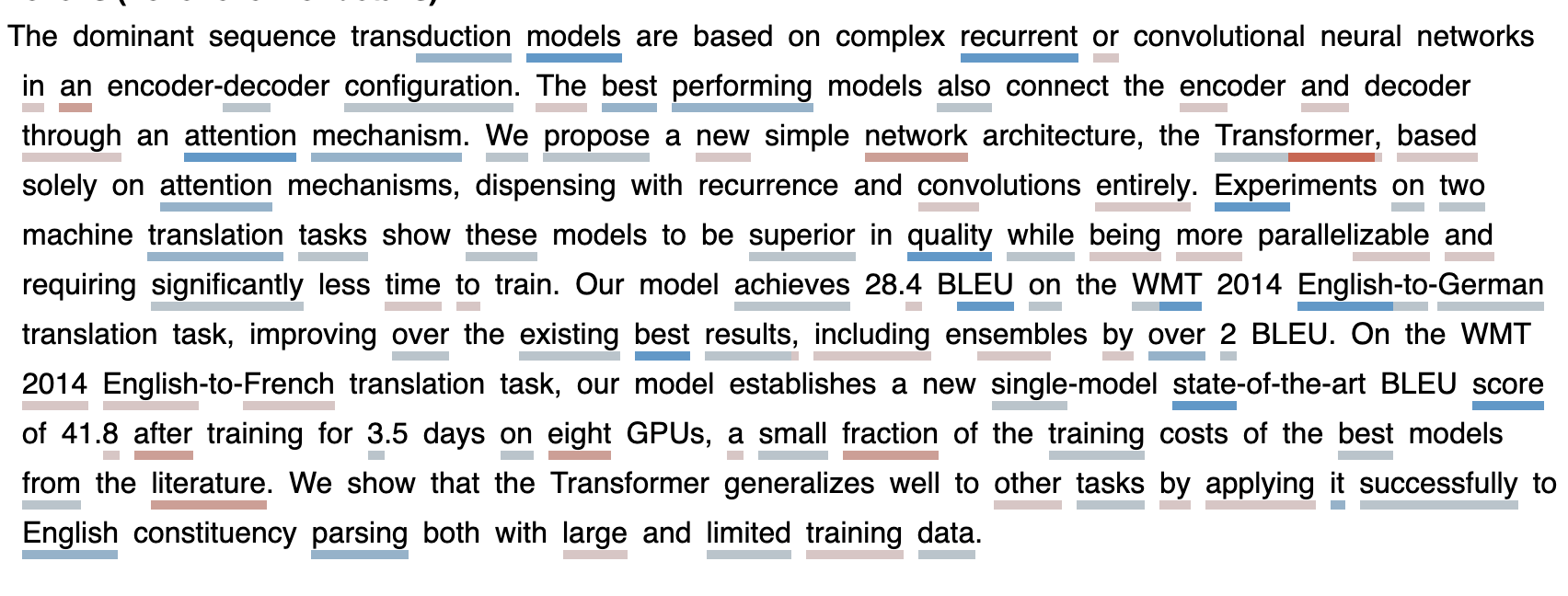}
\caption{\mone{GPT-2} vs \mtwo{GPT-Arxiv-nlp} on an abstract of an NLP paper. }
    \label{fig:arxiv}
\end{figure}

\begin{figure}[th]
    \centering
    \includegraphics[width=\columnwidth]{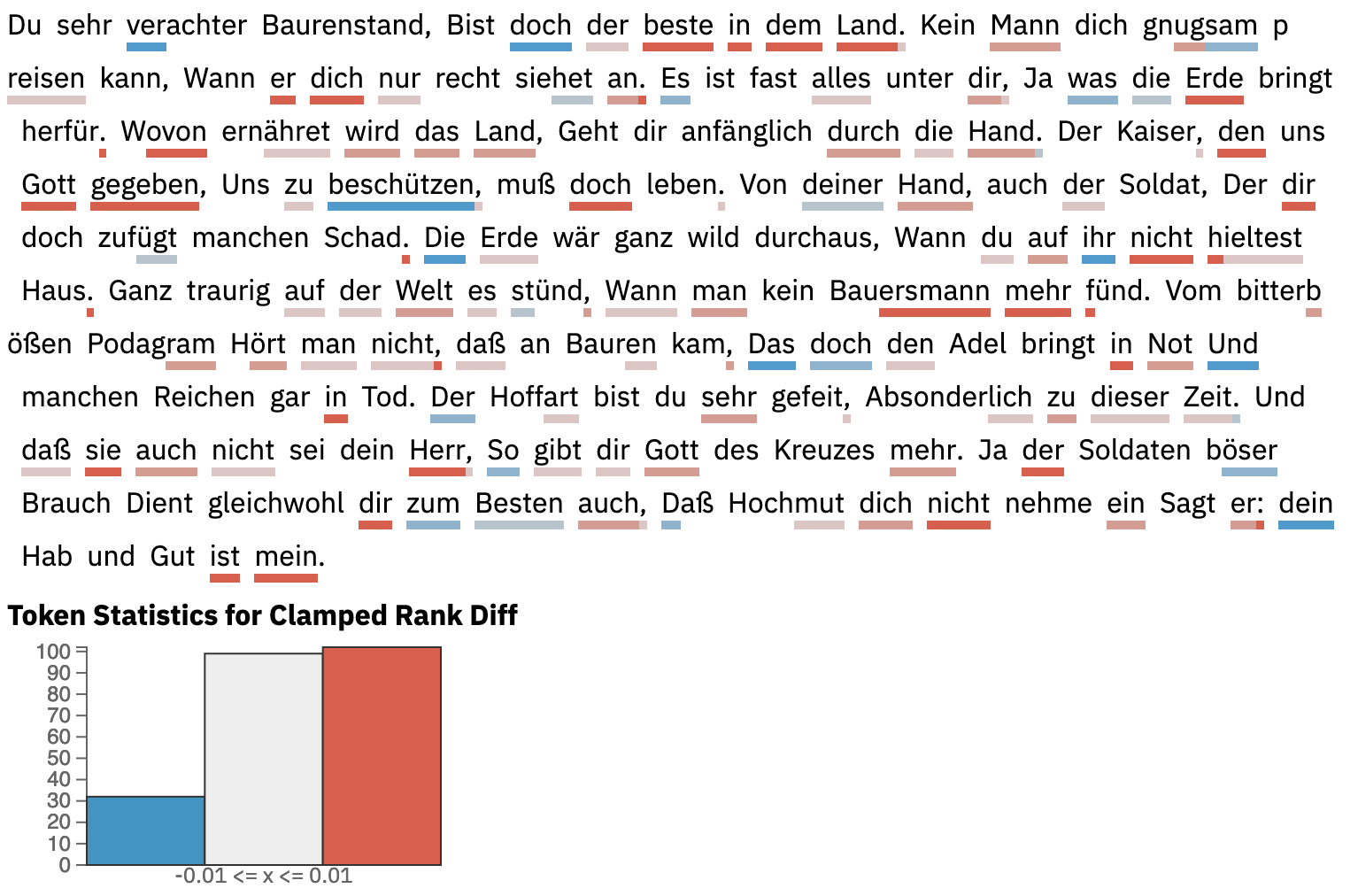}
\caption{\mone{GPT2-German} vs \mtwo{GPT2-German-Faust} on a snippet from the 1668 book ``Simplicius Simplicissimus'' using the Clamped Rank difference.}
    \label{fig:faust}
\end{figure}

\paragraph{In Domain}
In Figure~\ref{fig:arxiv}, we show a comparison between GPT-2 and GPT-2-ArXiv-NLP on an abstract of an NLP paper, highlighting the probability difference. As expected, NLP-specific terms (WMT BLEU, model, attention, etc.) tend to be more likely under the finetuned model. But, interestingly, the name of languages and Transformer are both more likely under the original model. This finding may warrant a deeper investigation for possible causes and whether this phenomenon persists across other contexts.

\paragraph{Out of Domain}
Out-of-domain tests can be useful for checking whether the finetuning process led to some transfer learning, or to test for catastrophic forgetting. In our case study, we compare GPT-2-German before and after finetuning on Goethe's Faust part I (1808) and II (1832). We hypothesized that the contemporary model would not be able to handle other works of literature from a similar time-period as well as the Faust-model, and thus tested on various snippets from books of the years 1200 to 1900. Our sample from the book Simplicius Simplicissimus (1668) (Figure~\ref{fig:faust}) is representative of the consistent finding that GPT-2-German performs better than the Faust variant. This could have many reasons --- the model may have overfit on the Faust-style of writing, the investigated periods of literature may differ too much, or they may differ too little from contemporary German.

\subsection{Finding dataset errors}
\label{sec:errors}

While not the original goal of \lmdiff, we observed that in some cases the outlier detection method could also be used to find outlier \textit{data} instead of examples where models differ significantly. 
One such example occurred when comparing GPT-ArXiv to GPT-2 on the BioLang dataset.
It appears that GPT-2 is much better at modeling repetitive, nonsensical character sequences which were thus surfaced through the algorithm (see Appendix~\ref{app:case-studies}).

\section{System Description}
\label{sec:design}

All comparisons in \lmdiff{} begin with three provided arguments: a dataset containing the interesting phases to analyze, and two comparable Transformer models. \lmdiff{} wraps Huggingface Transformers~\citep{wolf-etal-2020-transformers} and can use any of their pretrained autoregressive or masked language models from their model hub\footnote{\url{https://huggingface.co/models}} or a local directory. Two models are comparable if they use the same tokenization scheme and vocabulary. This is required such that a phrase passed to either of them will have an identical encoding, with special tokens added in the same locations.

\lmdiff{} then does the work of recording each model's predictions across the dataset into an \textit{AnalysisCache}. Each token in each phrase of the dataset is analyzed for its ``rank'' and ``probability''. We define a token's rank as the affinity of the LM to predict the token relative to all other tokens, where a rank of 1 indicates it is the most favorable token, and the probability is computed from a direct softmax of the token's logit. Other useful information is also stored, such as the top-10 tokens (and their probabilities) that would have been predicted in that token's spot. This information can then be compared to other caches and explored in the visual interface. The interface can also be used independently of cache files to compare models on individual inputs.

The modular design separating \textit{datasets}, \textit{models}, and their \textit{caches} makes it easy to compare the differences between many different models on distinct datasets. Once a cache has been made of a (model, dataset\_D) pair, it can be compared to any other cache of a (comparable\_model, dataset\_D) pair within seconds. More information is provided in Appendix~\ref{app:system}.

\paragraph{Adding models and datasets} It is easy to load additional models and datasets. First, ensure that the model can be loaded through the Huggingface \codeword{AutoModelWithLMHead} and \codeword{AutoTokenizer} function \codeword{from_pretrained(...)} which supports loading from a local directory.
The following script prepares two models and a dataset for comparison:

\begin{lstlisting}
python scripts/preprocess.py all \ 
    [OPTIONS] M1 M2 DATASET \
    --output-dir OUT
- M1 = Path (or name) of HF model 1
- M2 = Path (or name) of HF model 2
- DATASET = Path to dataset.txt
- OUT = Where to store outputs
\end{lstlisting}

\noindent The output configuration directory \codeword{OUT} can be passed directly to the \lmdiff~server and interface which will automatically load the new data:

\begin{lstlisting}
python backend/server/main.py \
    --config DIR
- DIR = Contains preprocessed outputs
\end{lstlisting}

\noindent The interface works equally well to compare two models on individual examples without a preprocessed cache:

\begin{lstlisting}
python backend/server/main.py \
    --m1 MODEL1 --m2 MODEL2
\end{lstlisting}

\section{Discussion and Conclusion}
We presented \lmdiff, a tool to visually inspect qualitative differences between language models based on output distributions. We show in several use cases how finding specific text snippets and analyzing them token-by-token can lead to interesting hypotheses. 

We emphasize that \lmdiff{} by itself does not provide any definite answers to these hypotheses by itself -- it cannot, for example, show which model is generally better at a given task. To answer these kind of questions, statistical analysis is required. 

A design limitation of \lmdiff{} is that it relies on compatible models. Because the tool is based on per-token model outputs and apples-to-apples comparisons of distributions, only models that use the same tokenization scheme and vocabulary can be compared in the instance view. In future work, we will work toward extending the compatibility by introducing additional tokenization-independent measures and visualizations. 

Another extension of \lmdiff{} may probe for memorized training examples and personal information using methods proposed by \citet{carlini2020extracting}. As shown in Sections~\ref{sec:gltr} and \ref{sec:errors}, we can already identify text that was generated by a model and leverage patterns that a model has learned. Adding support to filter a corpus by measures in addition to finding outliers may help with the analysis of potentially memorized examples.

\section{Acknowledgements}
We thank Ankur Parikh and Ian Tenney for helpful comments on an earlier draft of this paper.
This work was supported by the MIT-IBM Watson AI Lab. This work has been developed in part during the BigScience Summer of Language Models 2021. 

\bibliography{anthology,custom}
\bibliographystyle{acl_natbib}

\appendix

\section{Additional Case Studies}
\label{app:case-studies}

\begin{figure}[b]
    \centering
    \includegraphics[width=\columnwidth]{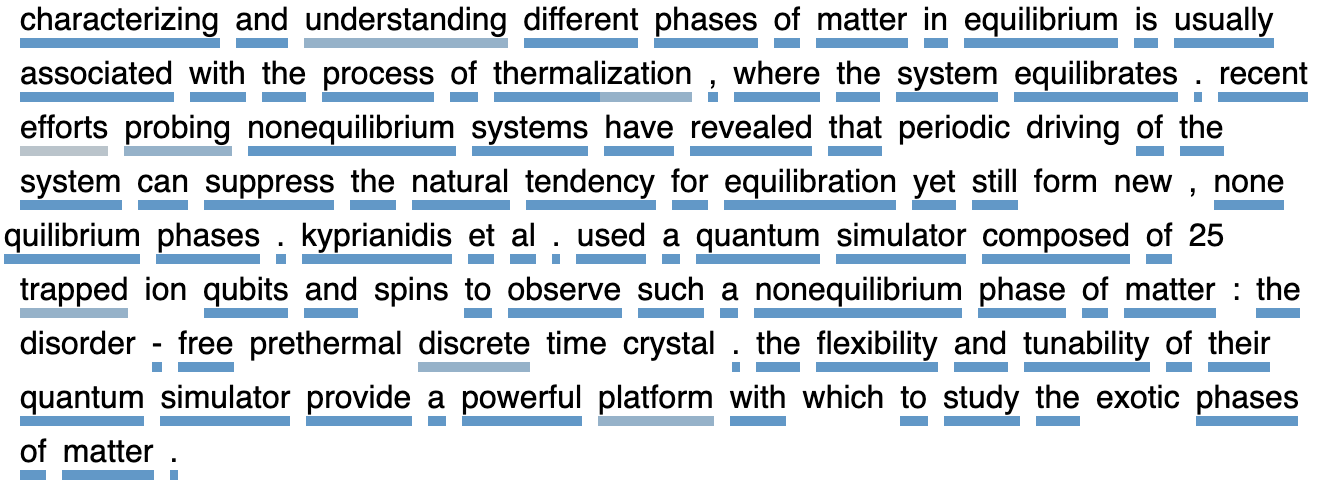}
\caption{\mone{DistilBERT-SST} vs \mtwo{DistilBERT} on a scientific abstract.}
    \label{fig:forgetting}
\end{figure}

\subsection{Masked LMs break when fine-tuning on different tasks}
When finetuning an autoregressive language model, the output representations are preserved since downstream tasks often make use of the language modeling objective. This is different for masked language models like BERT. Typically, the contextual embeddings are combined with a new untrained head and thus, the language modeling is ignored during finetuning. We demonstrate this in Figure~\ref{fig:forgetting} where we compare DistilBERT (blue) and DistilBERT-SST (red) on a recent abstract published in Science. DistilBERT performs much better, having a significantly higher probability for almost every token in the text. Since the finetuned model started with the same parameters, this is a particular instance of catastrophic forgetting~\citep{mccloskey1989catastrophic}. 
While this case is somewhat obvious, \lmdiff{} can help identify domains that are potentially more affected by this phenomenon even for cases in which the language modeling objective is not abandoned. 

\subsection{Data Outliers}
\begin{figure*}[t]
    \centering
    \includegraphics[width=\textwidth]{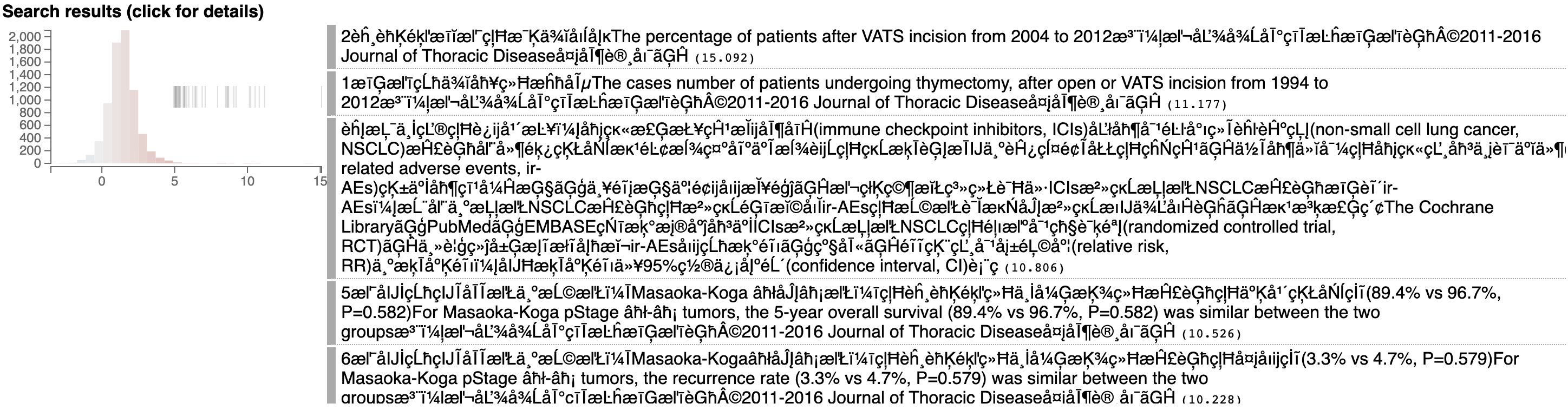}
\caption{BioLang with \mone{GPT-2} vs the \mtwo{GPT-2-ArXiv}. GPT-2 is much better at modeling repeated patterns which helps identify malformed examples.}
    \label{fig:dataset-errors}
\end{figure*}

We show one example of a data outlier, described in Section~\ref{sec:errors}, in Figure~\ref{fig:dataset-errors}. The top-ranked examples in the corpus all have severe encoding errors and those examples should be removed from the corpus.

\begin{figure}[t]
    \centering
    \includegraphics[width=\linewidth]{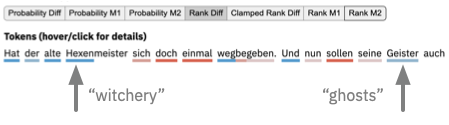}
    \caption{Magic characters more likely under GPT2-German-Faust}
    \label{fig:faust1}
\end{figure}

\begin{figure}[t]
    \centering
    \includegraphics[width=\linewidth]{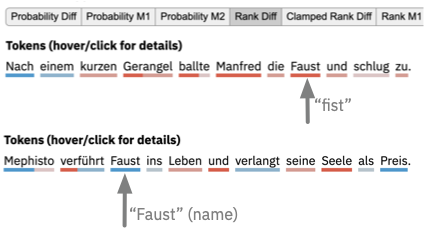}
    \caption{Tokens can be more likely under different models depending on contexts.}
    \label{fig:faust2}
\end{figure}

\begin{figure*}[htb]
    \centering
    \includegraphics[width=.95\linewidth]{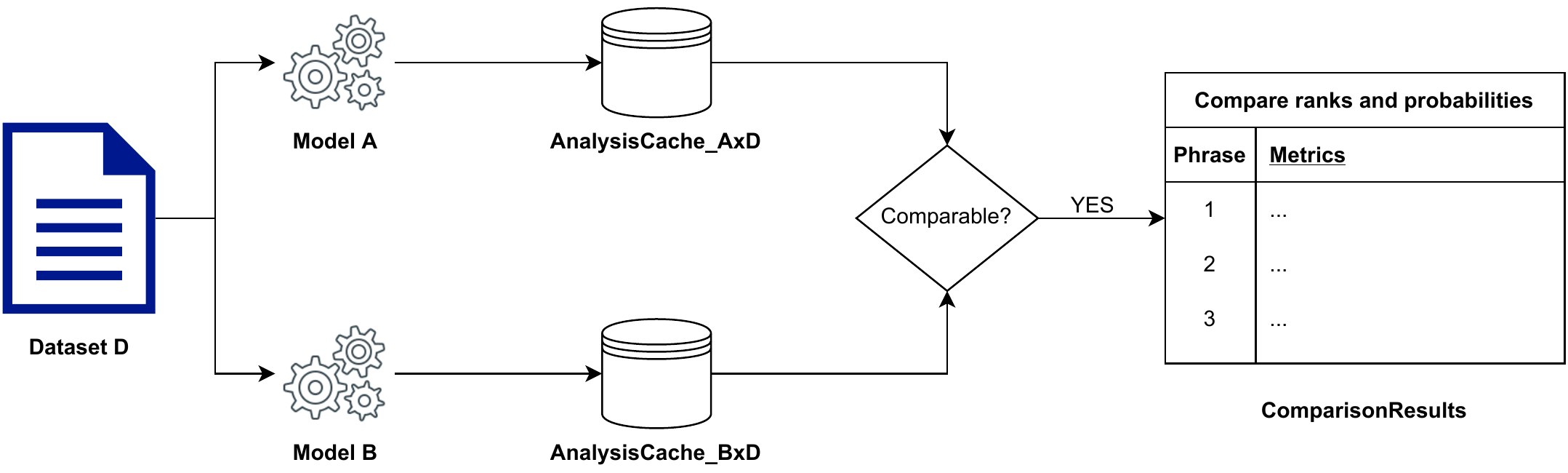}
\caption{System diagram of the \lmdiff{} backend. }
    \label{fig:system-design}
\end{figure*}

\subsection{Language specific to finetuned model}
The comparison of GPT2-German and GPT2-German-Faust (see Section~\ref{sec:domain}) also revealed more patterns that indicate that the fine-tuning of the model might have been successful. \autoref{fig:faust1} shows an example where tokens related to the core text of the Faust text are more likely under the fine-tuned model than the wild-type GPT2-German. Tokens like ``Hexe'' (witch)  or ``Geister'' (ghosts) are core characters in the Faust text. 

Another interesting observation is that even the same tokens in different contexts can be more likely under different models. The token ``Faust'' can refer to the name of the main character in the story or be the common German translation for ``fist''. \autoref{fig:faust2} shows how the word is more likely under the general language model if embedded in a fighting context versus being embedded in a one-sentence summary of the Faust story. 

\section{System diagram for corpus analyses}
\label{app:system}

Figure~\ref{fig:system-design} describes how \lmdiff{} identifies compatibility between models and precomputed corpora. The \textbf{Dataset} is a text file where each new line contains a phrase to analyze. It also contains a YAML header containing necessary information like its name and a unique hash of the contents. This \textit{dataset} is processed by different Huggingface Transformer \textbf{Models} that receive the contents of the dataset as input and make predictions at every token. The tokenizations and predictions for each of the phrases are stored in the \textbf{AnalysisCache}, which takes the form of an HDF5 file. Finally, any two \textit{AnalysisCaches} can be checked for comparability. If they are comparable, the difference between them can be summarized in a \textbf{ComparisonResults} table and presented through the aforementioned interface for inspection and exploration by the user.

\end{document}